%% file: main.tex
\if false
\documentclass[journal,twoside,web]{ieeecolor}
\usepackage{generic}
\usepackage{cite}
\usepackage{amsmath,amssymb,amsfonts}
\usepackage{algorithmic}
\usepackage{graphicx}
\usepackage{booktabs}
\usepackage{algorithm,algorithmic}
\usepackage{hyperref}
\hypersetup{hidelinks}
\usepackage{textcomp}
\usepackage{multirow}
\def\BibTeX{{\rm B\kern-.05em{\sc i\kern-.025em b}\kern-.08em
    T\kern-.1667em\lower.7ex\hbox{E}\kern-.125emX}}
\fi

\documentclass[journal]{IEEEtran}
\usepackage{cite}
\usepackage{graphicx}
\usepackage{algorithm,algorithmic}
\usepackage{tabularx}
\usepackage{booktabs}
\usepackage{multirow}
\usepackage{multicol}
\usepackage{amssymb}
\usepackage{amsmath}
\usepackage{authblk}
\usepackage[pagebackref=false,breaklinks=true,letterpaper=true,colorlinks,bookmarks=false]{hyperref}
\DeclareUnicodeCharacter{2212}{-}
\usepackage{algorithmic}
\usepackage{textcomp}
\usepackage{color}
\usepackage{multirow}
\usepackage[table,xcdraw]{xcolor}

\usepackage{silence}
\input{defs}

\begin{document}
\title{ECC-PolypDet: Enhanced CenterNet with Contrastive Learning for Automatic Polyp Detection}
\author{Yuncheng Jiang*, Zixun Zhang*, Yiwen Hu*, Guanbin Li, Xiang Wan, Song Wu, Shuguang Cui \IEEEmembership{Fellow, IEEE}, Silin Huang$^{\#}$,  Zhen Li$^{\#}$, \IEEEmembership{Member, IEEE}
\thanks{This work was supported by the Basic Research Project No. HZQB-KCZYZ-2021067 of Hetao Shenzhen HK S$\&$T Cooperation Zone, by Shenzhen-Hong Kong Joint Funding No. SGDX20211123112401002, by Shenzhen General Program No. JCYJ20220530143600001, by NSFC with Grant No. 62293482, by Shenzhen Outstanding Talents Training Fund, by Guangdong Research Project No. 2017ZT07X152 and No. 2019CX01X104, by the Guangdong Provincial Key Laboratory of Future Networks of Intelligence (Grant No. 2022B1212010001), by the Guangdong Provincial Key Laboratory of Big Data Computing, The Chinese University of Hong Kong, Shenzhen, by the NSFC 61931024$\&$81922046, by the Shenzhen Key Laboratory of Big Data and Artificial Intelligence (Grant No. ZDSYS201707251409055), and the Key Area R$\&$D Program of Guangdong Province with grant No. 2018B030338001, by zelixir biotechnology company Fund, by Tencent Open Fund.}
\thanks{*Yuncheng Jiang, Zixun Zhang, and Yiwen Hu have equal contributions to this work. $^{\#}$Silin Huang and Zhen Li are the equal corresponding authors.}
\thanks{Yuncheng Jiang and Zixun Zhang are with the Future Network of Intelligence Institute (FNii) \& the School of Science and Engineering (SSE), the Chinese University of Hong Kong, Shenzhen, and also with the Shenzhen Research Institute of Big Data (SRIBD) (email: \{yunchengjiang, zixunzhang\}@link.cuhk.edu.cn).}
\thanks{Shuguang Cui and Zhen Li are with the School of Science and Engineering (SSE) \& the Future Network of Intelligence Institute (FNii), the Chinese University of Hong Kong, Shenzhen (email:\{shuguangcui, lizhen\}@cuhk.edu.cn).}
\thanks{Yiwen Hu is with the School of Science and Engineering (SSE), the Chinese University of Hong Kong, Shenzhen, and also with South China Hospital of Shenzhen University (email: yiwenhu1@link.cuhk.edu.cn).}
\thanks{Guanbin Li is with the School of Data and Computer Science, Sun Yat-sen University (email: liguanbin@mail.sysu.edu.cn).}
\thanks{Xiang Wan is with the Shenzhen Research Institute of Big Data (email: wanxiang@sribd.cn).}
\thanks{Silin Huang and Song Wu are with South China Hospital of Shenzhen University (email: huangsilin214@163.com, wusong@szu.edu.cn)}
}

\maketitle

\begin{abstract}
Accurate polyp detection is critical for early colorectal cancer diagnosis. Although remarkable progress has been achieved in recent years, the complex colon environment and concealed polyps with unclear boundaries still pose severe challenges in this area. Existing methods either involve computationally expensive context aggregation or lack prior modeling of polyps, resulting in poor performance in challenging cases. In this paper, we propose the Enhanced CenterNet with Contrastive Learning (ECC-PolypDet), a two-stage training \& end-to-end inference framework that leverages images and bounding box annotations to train a general model and fine-tune it based on the inference score to obtain a final robust model. Specifically, we conduct Box-assisted Contrastive Learning (BCL) during training to minimize the intra-class difference and maximize the inter-class difference between foreground polyps and backgrounds, enabling our model to capture concealed polyps. Moreover, to enhance the recognition of small polyps, we design the Semantic Flow-guided Feature Pyramid Network (SFFPN) to aggregate multi-scale features and the Heatmap Propagation (HP) module to boost the model's attention on polyp targets. In the fine-tuning stage, we introduce the IoU-guided Sample Re-weighting (ISR) mechanism to prioritize hard samples by adaptively adjusting the loss weight for each sample during fine-tuning. Extensive experiments on six large-scale colonoscopy datasets demonstrate the superiority of our model compared with previous state-of-the-art detectors.
\end{abstract}

\begin{figure}[t]
    \centering
    \includegraphics[width=0.9\linewidth]{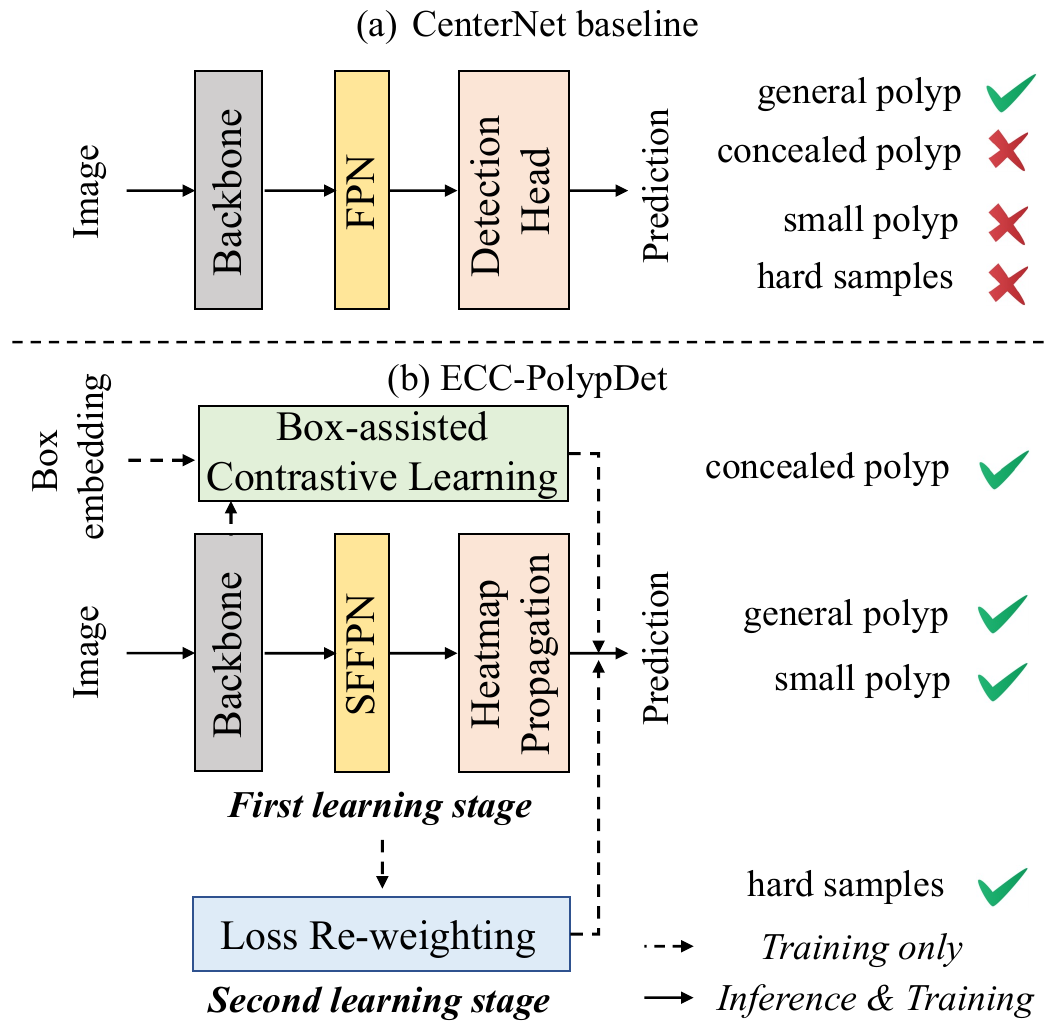}
    \caption{\textbf{Illustration of the pipeline of our ECC-PolypDet}. We add a supervised contrastive learning branch to increase the model's recognition capability of concealed polyps. We further modify the feature pyramid network (FPN) and detection head structure to capture small object features. Finally, we fine-tune the hard samples via a loss re-weighting method. During inference, our model follows an end-to-end manner. The modules that the dashed line flows through will be removed.
    }
    \label{fig:teaser}
\end{figure}

\section{Introduction}
\label{sec:introduction}

\IEEEPARstart{C}{olorectal} cancer (CRC) is a critical public health issue, with it being the third most frequently diagnosed cancer and the second leading cause of cancer-related deaths worldwide~\cite{CRC_review}. More than 80\% of colorectal cancers arise from polyps~\cite{survival_trend}, making the identification and removal of malignant polyps crucial in reducing CRC-based mortality rates~\cite{cancer_2023}. Therefore, colonoscopy is regarded as the golden standard technique for early polyp screening. However, the traditional clinical colonoscopy diagnosis suffers a high miss rate (as much as 27\%)~\cite{polyp_miss_rate} due to the irregular operation and negligence of endoscopists after long duty. Fortunately, the development of Convolutional Neural Networks (CNNs) and Transformers has led to the creation of numerous detection models that have shown remarkable progress. Automatic polyp detection systems have been developed to assist endoscopists and minimize the risk of misdiagnosis~\cite{miccai2015comparative}.

Despite the remarkable progress made in the development of detection models, accurate and reliable polyp detection remains a challenging task due to three primary challenges: 
1) \textbf{Concealed polyp.} 
Most polyps in the colonoscopy videos exhibit a similar appearance to the surrounding colorectal tissue, especially in low-light conditions and cases of inadequate bowel preparation. This similarity poses challenges even for experienced endoscopists in accurately identifying these polyps. Similarly, conventional CNN models struggle to accurately distinguish foreground polyps from irrelevant backgrounds. Previous methods of polyp detection attempted to collaborate multi-frame temporal information to align the features cross frames~\cite{stft}. However, the subtle visual appearance contrasts between consecutive frames are insufficient in extracting the discriminative features required for accurate polyp identification.
2) \textbf{Small and flat polyps.} 
The majority of polyp regions are relatively small compared to the image size. Our analysis (\ref{ssec:dataset}) shows that the size of most polyps in our datasets is less than 0.1\% of the image size. It poses two significant challenges for existing detection methods. Firstly, there is a severe region imbalance between the foreground (polyp) and background (colorectal wall), which can cause the polyp region to be overwhelmed by the large background and lead the model to overfit irrelevant information. Secondly, small polyps are difficult to be accurately labeled, which further exacerbates the unclear boundary issue.
3) \textbf{Severe imbalance of hard samples.}
The imbalance distribution of the easy and hard samples is a long-standing problem in the object detection task. Hard samples refer to images containing polyps or internal artifacts that are difficult for the model to focus on and learn from during training, resulting in low inference scores. These samples may contribute to low training loss, potentially causing the model to converge to a local minimum.

In this paper, we propose the ECC-PolypDet, a polyp detection model designed to tackle the challenges mentioned above. Fig.~\ref{fig:teaser} shows the basic pipeline of our model. Specifically, Considering the high homogeneity of polyp targets, we design a bounding box-assisted contrastive learning (BCL) module to train ECC-PolypDet. By using the bounding box annotations, we divide the input image into the foreground and background classes and contrastively optimize the distance of features between the two classes to minimize the intra-class distance while maximizing the inter-class distance. Furthermore, we employ the semantic flow-guided FPN (SFFPN) to align spatial information between multi-scale features and leverage a heatmap propagation (HP) module to progressively capture contextual information during the detection stage. Finally, considering the severe imbalance of hard samples, we introduce an effective IoU-guided sample re-weighting (ISR) strategy to optimize our ECC-PolypDet. This strategy adaptively adjusts the loss of each sample during the second learning stage according to their inference IoU scores, resulting in a more robust and generalizable detection model.

In summary, our contributions are listed as follows: 

\begin{itemize}
    \item We propose the ECC-PolypDet, a two-stage training and end-to-end inference framework for accurate and robust polyp detection. It comprises a first learning stage that focuses on general sample learning and a second learning stage dedicated to understanding hard samples.
    \item In the first learning stage, to better distinguish the features between polyp and background, we design a bounding box-assisted contrastive learning framework to jointly train the detection network. \textit{We are the first to leverage bounding box information for supervised contrastive learning applied in the polyp detection task.}
    \item To further enrich the feature of small polyps, we first employ the semantic flow-guided FPN to effectively aggregate multi-scale features from the backbone with minimal information loss. Then, we incorporate the heatmap propagation module into the CenterNet architecture to progressively refine the small features. In the second learning stage, we design an IoU-guided sample re-weighting strategy to optimize the network with adaptive weights that emphasize more on hard samples. 
    \item We conduct extensive experiments on six datasets (\textit{i.e.} SUN Colonoscopy Video Database~\cite{suncolon,sunwebsite}, LDPolypVideo~\cite{ldpolypvideo}, CVC-VideoClinicDB~\cite{cvc_1,cvc_2}, PolypGen~\cite{ali2023multi}, LHR Database-L/S) and establish state-of-the-art new performances, which demonstrate the superiority of our proposed framework.
\end{itemize}

\begin{figure*}[t]
    \centering
    \includegraphics[width=0.9\linewidth]{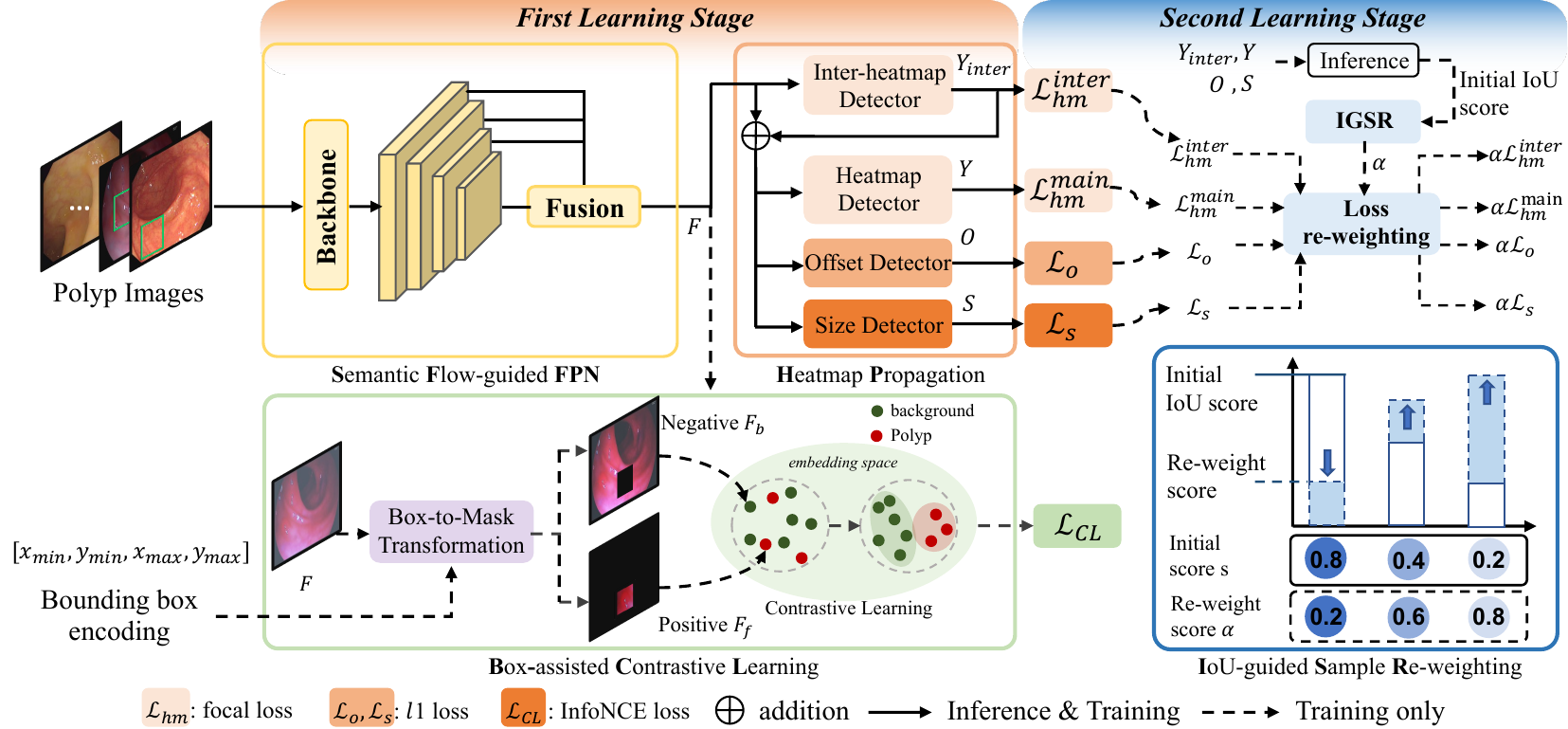}
    \caption{\textbf{Detailed illustration of our ECC-PolypDet framework}. It consists of the Semantic Flow-guided FPN (SFFPN), the CenterNet with a Heatmap Propagation (HP) module, a Box-assisted Contrastive Learning (BCL) module, and an IoU-guided Sample Re-weighting (ISR) module. Our ECC-PolypDet is jointly trained with detection loss and contrastive loss. After the first learning stage, the model is finetuned with adaptive sample importance weight processed by ISR.
    During inference, BCL and ISR modules will be removed. Only SFFPN and HP modules are adopted for prediction. The algorithm of our pipeline is presented in Alg.~\ref{alg:system}. }
    \label{fig:framework}
\end{figure*}

\section{Related Works}
\label{sec:relate}
In this section, we will introduce the progress of automatic colonoscopy polyp detection algorithms from two categories: \textbf{hand-crafted} and \textbf{deep learning}.

\subsection{Polyp Detection based on hand-crafted feature}
Automated polyp detection as a computer-aided clinical endoscopic diagnostic technique has been an active research topic for decades. As in the early stage, the majority of the methods involved extracting hand-crafted features from color, shape, and texture based on low-level image processing techniques to identify the candidate polyp regions. Subsequently, classifiers like Bayes or SVM were employed for diagnosis. Karkanis~\textit{et al.}~\cite{karkanis2003computer} conducted wavelet decomposition to detect whether the collected image contains polyps. ~\cite{Texture_Based} used the polyp's shape and appearance features as descriptors to guide the classification and localization. \cite{iwahori2013automatic} utilized Hessian filters to capture the polyp boundaries. Karkanis~\textit{et al.}~\cite{Local_Binary} leveraged the color wavelet features combined with a sliding window for polyp detection. Tajbakhsh~\textit{et al.}~\cite{Shape_Context} exploited both edge detection and feature extraction to boost detection accuracy. Zhu~\textit{et al.}~\cite{zhu2010improved} analyzed curvatures of detected boundaries to find the polyps. Ren~\textit{et al.}~\cite{ren2016high} proposed to combine the shape index and the multi-scale enhancement filter by Gaussian smooth distance field to generate the candidate polyp. However, those methods achieved poor performance and significantly false-positive due to the inaccurate hand-crafted features given the variant conditions in colonoscopy~\cite{miccai2015comparative}.

\subsection{Polyp Detection based on deep learning algorithm}
Deep learning-based methods have surpassed the capabilities of traditional hand-crafted features in terms of feature representation. Consequently, numerous automatic polyp detection approaches have emerged and have been applied to computer-aided diagnosis. In the early stage, colonoscopy polyp detection is considered as object detection, where those methods are generally grouped into two categories: "two-stage detection" and "one-stage detection." The former defines detection as a "coarse-to-fine" process, such as Faster R-CNN~\cite{faster_rcnn}. Those methods first look over region proposals globally in the image and regress the bounding box in each of the proposals. On the other hand, the latter defines detection as "completion in one step", such as the YOLO series~\cite{yolox, yolov3}. However, these methods suffer from inflexibility due to the extensive use of anchor boxes, which leads to a limited range of scales and aspect ratios. Afterward, popular detection technologies gradually abandoned anchors and formulated the object detection problem as a key-point detection problem. Specifically, represented by centernet~\cite{centernet}, the bounding boxes are predicted by regressing the distance from the keypoint to the boundaries. Besides, Sparse R-CNN~\cite{sparsercnn} proposes regressing the object box only by a small set of sparse learnable proposals. More recently, the powerful attention mechanism supports the emergence of DEtection TRansformers (DETR)~\cite{deformabledetr,dino}. DETR and its variants define detection as set prediction, which significantly simplifies the detection pipeline and achieves dominant performance on the detection task. Nevertheless, the computational cost is high for DETR in training and inference, which brings obstacles to real-time clinical application.

In addition to the conventional detection methods, many studies have been dedicated to developing algorithms tailored specifically for colonoscopy images, aiming to achieve a balance between high accuracy and real-time performance.~\cite{pacal2020comprehensive}. Debesh~\textit{et al.}~\cite{colono_seg} designed a real-time polyp detection and segmentation system, achieving nearly 180 frames per second (FPS) inference speed. SSL-CPCD~\cite{xu2023ssl} investigated the generalization issues in colonoscopic image analysis and introduced a novel self-supervised learning method with instance-group discrimination to improve model performance. STFT~\cite{stft} proposed a multi-frame collaborative framework to adaptively mine spatiotemporal correlations with carefully designed spatial alignment and temporal aggregation. Gong~\textit{et al.}~\cite{gong2023frcnn} introduced a two-stage detection model and applied attention awareness and context fusion to detect colon polyps. Notably, the YOLO algorithms have attracted attention due to their desirable features of fast inference and low computational burden. Pacal~\cite{pacal2021robust} first proposed several new structures on the YOLOv4 algorithm and achieved real-time polyp detection on the CVC-ColonDB challenge. Consequently, they designed a fast detection algorithm based on the YOLOv4 framework by integrating negative sample features~\cite{pacal2022efficient}. Meanwhile, the self-attention mechanism was first introduced in YOVOv5 by~\cite{wan2021polyp} and obtained SOTA results on Kvasir-SEG. Karaman~\textit{et al.}~\cite{karaman2023hyper} first demonstrated the importance of hyper-parameter optimization in polyp detection. They proved that combining optimization algorithms with a real-time detection framework, such as scaled-YOLOv4, is an efficient manner. They further improved the performance on SUN and PICCOLO polyp datasets by integrating YOLOv5 algorithms~\cite{karaman2023robust}. Lee~\textit{et al.}~\cite{lee2022improvement} proposed a real-time polyp detection system based on YOLOv4. In this system, a multi-scale mesh was used to detect small polyps. Despite the notable progress, there still exists a significant disparity between these methods and real-world clinical applications. Previous methods either lack specialized designs for addressing hard samples or rely on complex modules that result in computationally intensive inference processes. Inspired by this, we propose designs to deal with challenging polyp samples. Notably, these designs bring no computational costs to inference. Overall, our model follows a two-stage training but an end-to-end inference paradigm.

\section{Methodology}
\label{sec:methodology}
In this section, we will dip into the core of our method. We first outline the whole pipeline of our architecture and training protocol in section~\ref{ssec:framework}. Then we will describe the details of each component, including the semantic flow-guided FPN (section~\ref{ssec:fgfpn}), the heatmap propagation (section~\ref{ssec:cs}), the box-assisted contrastive learning (section~\ref{ssec:bacl}), and IoU-guided sample re-weighting mechanism (section~\ref{ssec:ahsm}). 
\footnote{This study with included experimental procedures and data were approved by the South China Hospital of Shenzhen University on May 12, 2023 (approval no.HNLS20230512001-A)}

\begin{figure}[t]
    \centering
    \includegraphics[width=\linewidth]{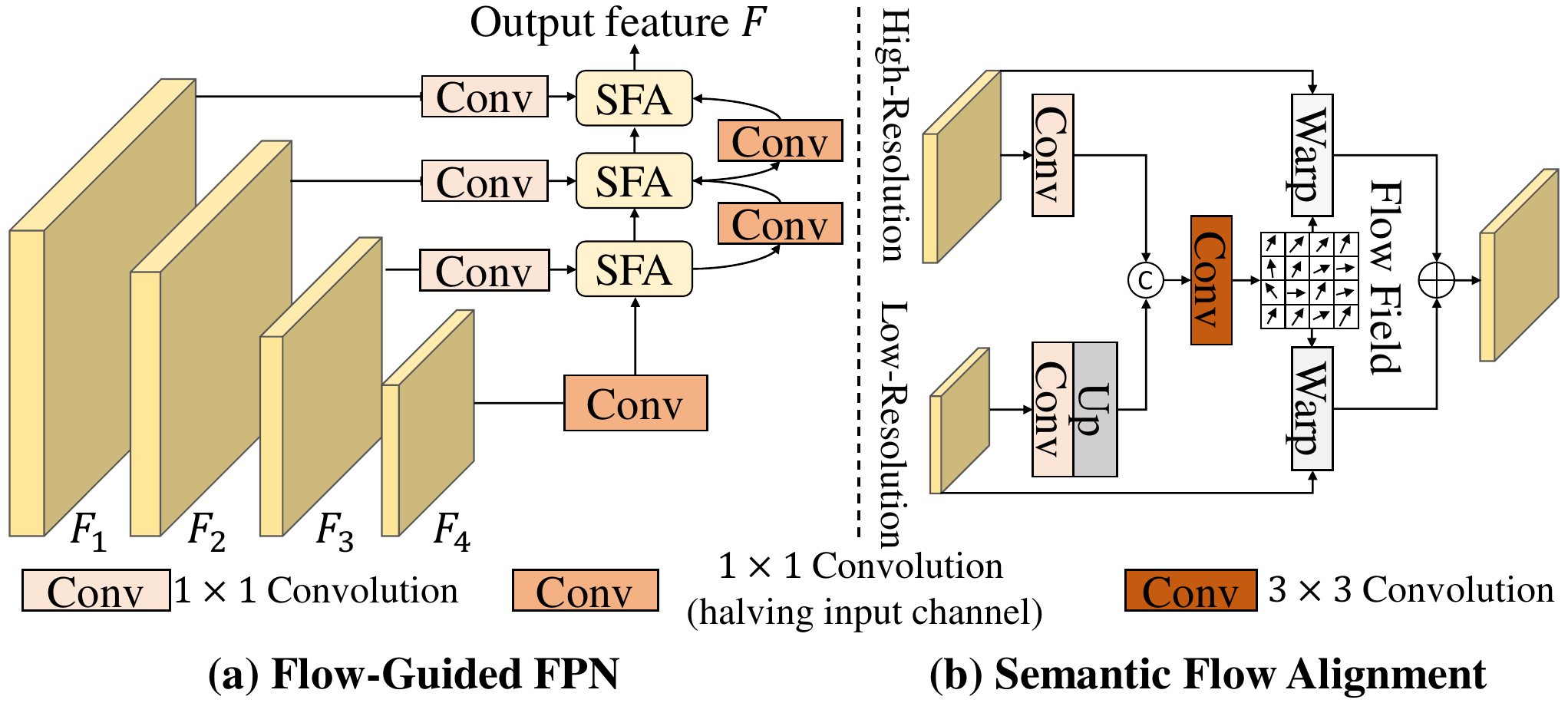}
    \caption{(a) Overview of the Semantic Flow-guided Feature Pyramid Network (SFFPN). (b) Details of the semantic flow alignment module (SFA). SFA learns semantic flow from high and low resolution features, and SFFPN fuses different scales of features to a high-resolution feature.}
    \label{fig:FGFPN}
\end{figure}

\subsection{Overall Framework}
\label{ssec:framework}

The overall pipeline of our proposed ECC-PolypDet framework is shown in Fig.~\ref{fig:framework}, which employs a two-stage training but an end-to-end inference manner.

Specifically, in the first learning stage, the information flow of the detector starts with an input image $I \in \mathbb{R}^{H \times W \times 3}$, which is first fed through the backbone network to extract multi-resolution features, $F_i \in \mathbb{R}^{\frac{H}{2^{i+1}} \times \frac{W}{2^{i+1}} \times C_i}, i \in \{1, 2, 3, 4\}$. Then, these features are fused via a Semantic Flow-guided Feature Pyramid Network (SFFPN) to resolve the misalignment caused by multiple downsampling of the backbone network. The fused feature $F$ undergoes processing by detection layers to generate heatmap, size, and offset features. To capture small targets, we recurrently enhance the heatmap layer several times as intermediate heatmap features and fuse them with the output heatmap features. Afterward, the fused features are used for box decoding. In addition, we introduce a Box-assisted Contrastive Learning (BCL) module to better distinguish polyps from the background. Specifically, we decouple the fused feature $F$ into foreground and background features according to the ground truth bounding box. These features are exploited to optimize the distance of intra/inter-class.

After the first learning stage, we observed a noticeable gap between the inference IoU score and training loss. Thus, a second learning stage with a simple yet effective IoU-guided sample re-weighting (ISR) mechanism is used to adjust the loss of each sample according to their inference IoU. Both the first/second learning stages are conducted in the training process. In the inference stage, we exclusively employ SFFPN to fuse features and predict object features in HP. Thus, the majority of computational overhead remains in the training process, ensuring the retention of high accuracy and speed in the inference stage.

\subsection{Semantic Flow-guided Feature Pyramid Network}
\label{ssec:fgfpn}
In CNN, the low-resolution feature in the deep layers is crucial for capturing global patterns, while the high-resolution features extracted from the shallow layers are essential to learning detailed information or small structures. To obtain a fine-grained representation for detection, traditional object detectors employ FPN to recover the downsampled features with the upsampling operation and gradually fuse the multi-scale features. However, the repeated downsampling and upsampling in the backbone and FPN leads to severe semantic misalignment between features of different scales, damaging the recognition of small objects~\cite{li2020semantic}. To this end, we propose the semantic flow-guided FPN to effectively aggregate multi-scale features, which lose minimal context information. The overall architecture of the SFFPN is shown in Fig.~\ref{fig:FGFPN} (a). At each stage $i$, The low-resolution feature $F_i$ is first compressed into the same channel with the adjacent high-resolution feature $F_{i-1}$ through $1\times1$ convolution and batch normalization (BN) layer. Then, two features are fused in the semantic flow-guided alignment module (SFA) to recover high-resolution and maintain context information. Fig.~\ref{fig:FGFPN} (b) shows the details of SFA. First, the low-resolution feature is upsampled to the same size as the high-resolution feature. Next, high and low-resolution features are concatenated to compute the semantic flow field via $3\times3$ convolution and BN layer, which represents the pixel offset between two feature scales. Then, a Warp function based on \cite{li2020semantic} is used to align two features according to the flow field. Finally, the warped features are summed as the output of the fused features, which encodes both rich context information and high-resolution features.
    
\subsection{Enhanced CenterNet with Heatmap Propagation}
\label{ssec:cs}

To achieve fast and accurate detection, we adopt the CenterNet, a simple and efficient anchor-free one-stage detector, as our baseline. In practice, CenterNet uses a Gaussian kernel to splat all ground truth object center points to heatmap $\tilde{Y}_{xy} = {\rm exp}(-\frac{(x-\tilde{p_x})^2+(y-\tilde{p_y})^2}{2\sigma_p^2} ), \tilde{Y} \in [0,1]^{W\times H}  $, where $ (\tilde{p_x}, \tilde{p_y}) $ is the coordinate of the kernel center and $ \sigma_p $ is the object size-adaptive standard deviation~\cite{centernet}. Each object is assigned a Gaussian kernel according to its size. However, due to the severe class imbalance between foreground and background, small polyps are prone to be overwhelmed by large backgrounds during training. Our ablation experiments also show that CenterNet lacks the ability to capture small polyps. 

Motivated by the Cascade R-CNN~\cite{cascade_rcnn}, we introduce a heatmap propagation (HP) module into CenterNet to gradually increase the attention of the network on small objects. Specifically, we insert a sequence of intermediate heatmap layers before the original prediction layers to enhance the heatmap features progressively. As shown in Fig.~\ref{fig:framework}, in the training stage, the fused feature $F$ is firstly processed by the intermediate heatmap layer, which is a sequence of convolution blocks with two $k \times k$ convolutions and outputs the intermediate heatmap $Y_{inter}$. Then, the intermediate heatmap is added to the original features $F$ using pixel-wise addition operation as the input for the next layer to refine the prediction features. Finally, all the intermediate heatmap, main-stream heatmap, size, and offset features are used to supervise the network. While in the inference stage, we use an ensemble of the intermediate and main-stream heatmaps for the bounding box prediction.

\begin{figure}[t]
    \centering
    \includegraphics[width=0.9\linewidth]{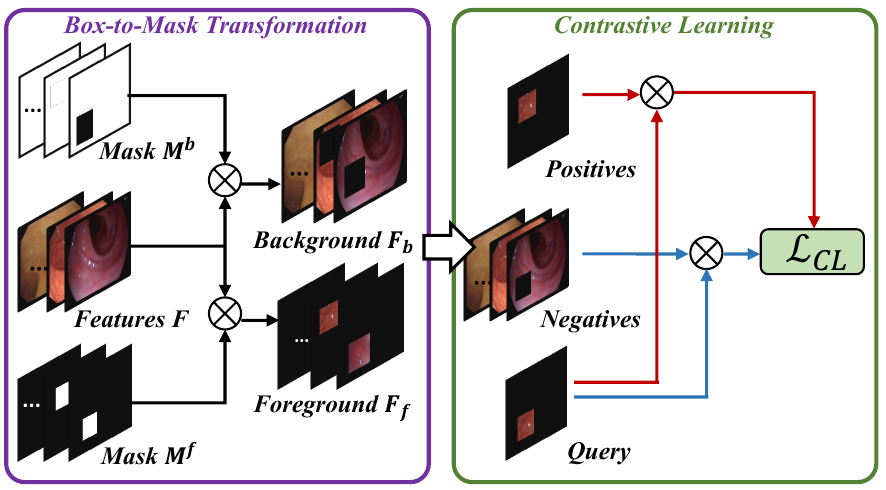}
    \caption{Process of box-assisted contrastive learning. 
    (BCL). In the Box-to-Mask Transformation stage, we use the bounding box annotation to generate the binary masks. The fused features are merged with the masks to get foreground and background features. 
    Next, in the contrastive learning stage, two random foreground features and all the background features are used to calculate the $L_{CL}$ (infoNCE loss).
    }
    \label{fig:BACL}
\end{figure}  

\subsection{Box-assisted Contrastive Learning}
\label{ssec:bacl}
Typically, polyp targets have a similar appearance to their surrounding tissues in terms of color and texture. Thus, designing a specialized mechanism to distinguish between foreground and background is beneficial for polyp detectors. Inspired by recent studies on supervised contrastive learning~\cite{scl,pixel}, propose a box-assisted contrastive learning module under the guidance of bounding box annotations.

The detailed process of our box-assisted contrastive learning module is shown in Fig.~\ref{fig:BACL}. Specifically, given fused feature maps $ F \in \mathbb{R}^{N \times \frac{W}{4} \times \frac{H}{4} \times C} $ associated with ground-truth bounding box $ G=\{ (x_i^{lt},y_i^{lt},x_i^{rb},y_i^{yb}), i=1,2, \cdots\} $. We first upsample the fused features to the original input size to obtain $F^{up}$. Then, we use the box annotations to generate the foreground binary masks $M^{f}$ where the polyp region is assigned with one while the background with zero, and the background binary masks $M^{b}$ in a similar way.

After that, we extract the feature embeddings of the foreground (polyp) $F_f$ and background $F_b$ using these masks. We filter $F^{up}$ by the binary masks on the spatial dimension ($\mathbb{R}^{N \times H \times W \times C} \rightarrow \mathbb{R}^{N \times 1 \times C}$) with masked average pooling (MAP) operation then normalize them to $[0,1]$:
\begin{equation}
\begin{aligned}
    F_f &= \text{norm}(\text{MAP}(F^{up}, M^{f})) \\
    F_b &= \text{norm}(\text{MAP}(F^{up}, M^{b})) \\
    \text{MAP}(F, M) &= \Sigma_{M_{i,j}=1}(F_{i,j})/\Sigma_{M_{i,j}=1}(M_{i,j})
\end{aligned}
\end{equation}
Next, for each foreground feature, we randomly select another foreground feature in the batch as the positive, while all the background features in the batch are taken as the negatives. Finally, we calculate the contrastive loss using infoNCE as:
\begin{equation}
    \mathcal{L}_{i}^{\rm NCE} = {-{\rm log}\frac{{\rm exp}(q_i \!\cdot\! i^+ / \tau)}{{\rm exp}(q_i \!\cdot\! i^+/\tau) \!+\! \sum_{i^- \!\in\! \mathcal{N}_i}{{\rm exp}(q_i \!\cdot\! i^- / \tau)}}} \\
\end{equation}
\begin{equation}
    \label{eq:ct_loss}
    \mathcal{L}_{CL} = \frac{1}{N} \sum_{i=1}^N \mathcal{L}_{i}^{NCE}
\end{equation}
\noindent where $q_i$ is the query, $i^+$ is randomly selected from the positives and $ \mathcal{N}_i$ denote embedding collections of the negatives.

\begin{figure}[t]
    \centering
    \includegraphics[width=0.75\linewidth]{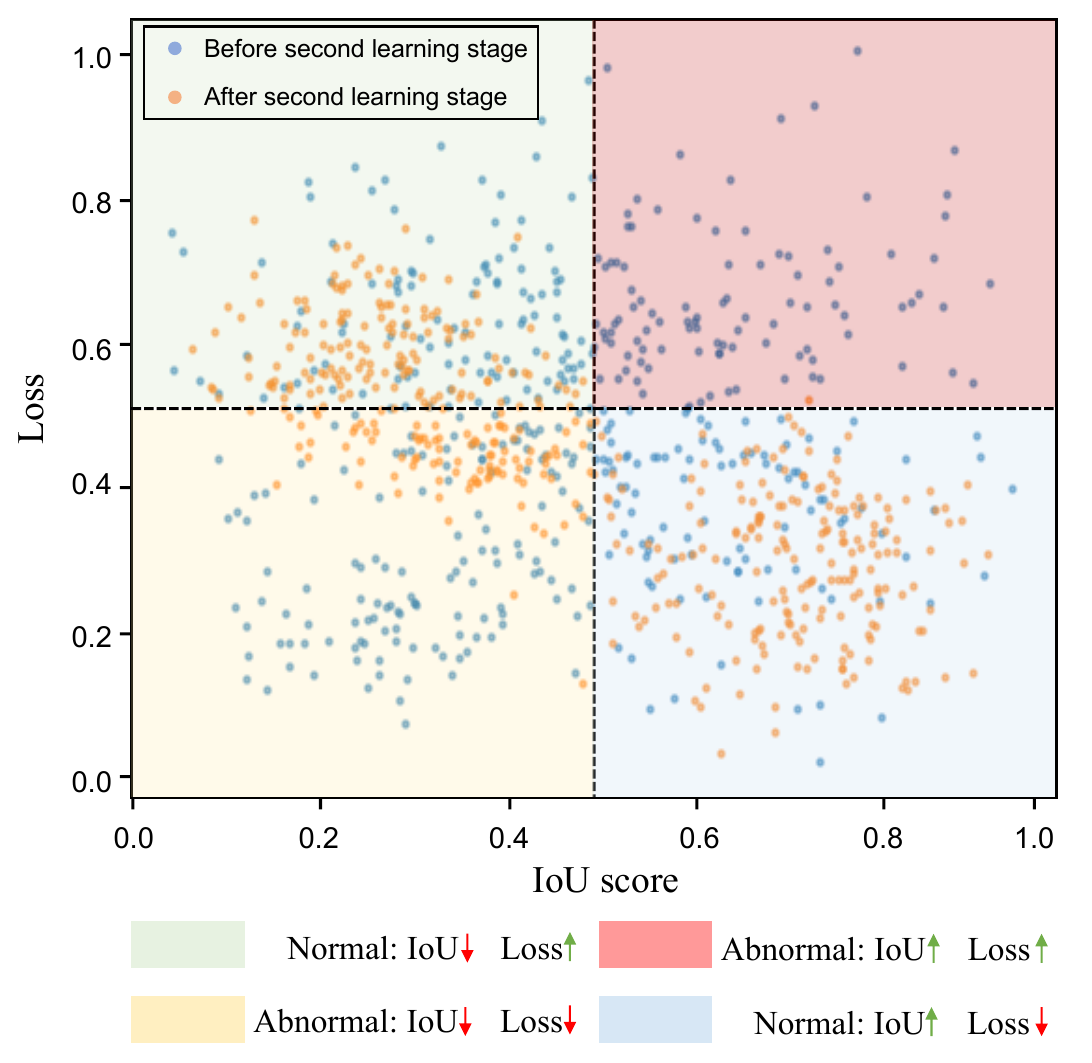}
    \caption{Visualization of the effect of adaptive hard mining strategy on the training loss of images. The point denotes the IoU-loss relationship. Before the fine-tuning stage, the points were scattered, and a large number of samples were distributed in abnormal areas. After fine-tuning, most samples were concentrated in normal areas.}
    \label{fig:hard_mining_result}
\end{figure}
\subsection{Loss Function}
\label{ssec:loss}

Following the standard definition of CenterNet, we use the focal loss with pixel-wise logistic regression as the detection loss to optimize the intermediate and main-stream heatmap:
\begin{equation}
    \label{eq:inter_hm}
    \mathcal{L}_{hm}^{inter}=-\frac{1}{N}\sum_{xy}\left\{\begin{matrix}
    (1-\tilde{Y}_{xy})^\alpha {\rm log}(\tilde{Y}_{xy}) & Y_{xy}=1 \\
    (1-Y^I_{xy})^\beta (\tilde{Y}_{xy} ){\rm log}(\tilde{Y}_{xy}) & \text{else}
    \end{matrix}\right.
\end{equation}
\begin{equation}
    \label{eq:hm}
    \mathcal{L}_{hm}^{main}=-\frac{1}{N}\sum_{xy}\left\{
    \begin{matrix}
    (1-\tilde{Y}_{xy})^\alpha {\rm log}(\tilde{Y}_{xy}) & Y_{xy}=1 \\
    (1-Y_{xy})^\beta (\tilde{Y}_{xy} ){\rm log}(\tilde{Y}_{xy}) & \text{else}
    \end{matrix}\right.
\end{equation}
To optimize the discretization error, the L1 loss is used to minimize the error of the predicted local offset ${{O}}\in \mathcal{R}^{H\times W\times 2} $ of each ground truth object center point $ \tilde{p} $ :
\begin{equation}
    \label{eq:offset}
    L_{o} = \frac{1}{N}\sum_{\tilde{p}} |{{O}}_{\tilde{p}} - \tilde{p} |  
\end{equation}
Finally, another L1 loss is set to optimize the size prediction of the object bounding box:
\begin{equation}
    \label{eq:size}
    \mathcal{L}_{s} = \frac{1}{N} \sum^N_{k=1}| {{S}}_{\tilde{p}k} -\tilde{s_{k}}| 
\end{equation}
where $ {{S}}_{\tilde{p}k} $ is the prediction of the size of k-th polyp and $ \tilde{s_{k}}  $ is the size of ground truth bounding box. 

Overall, the total training objective function is:
\begin{equation}
    \label{eq:totalloss}
    \mathcal{L} = \mathcal{L}_{hm}^{main} + \lambda_{inter}\mathcal{L}_{hm}^{inter} + \lambda_{o} \mathcal{L}_{o} + \lambda_{s} \mathcal{L}_{s} + \lambda_{CL} \mathcal{L}_{CL}
\end{equation}

\begin{algorithm}[t!]
    \caption{Training process of ECC-PolypDet}
    \label{alg:system}
    \begin{algorithmic}[1]
        \REQUIRE ~~\\ 
            Training images $\{X\}$; Ground truth bounding box embeddings $\{G\}$; Initial model weight $\Phi$; Optimizer $\mathcal{A}$; Max iteration $K$\\
        \ENSURE ~~\\ 
            Trained model weight $\Phi^{*}$\\
        \textbf{First learning stage:}
        \STATE $\Phi^0 := \Phi$;
        \FOR{\textit{i}=1 \textbf{to} $K$}
        \STATE Sample a mini-batch of $X_i$ and $G_i$ from $D$;
        \STATE Obtain fused features $F$, intermediate heatmap $Y_{inter}$, main heatmap $Y$, offset $O$ and size prediction $S$;
        \STATE Calculate the total loss Eq.~\ref{eq:totalloss} using $F$, $Y_{inter}$, $Y$, $O$, $S$ and $G_i$;
        \STATE Update model weight $\Phi^i := \mathcal{A}(\mathcal{L}, \Phi^{i-1})$;
        \ENDFOR\\
        
        \textbf{Hard sample mining:}
        \STATE Inference $\{X\}$ with $\Phi^{K}$ to obtain IoU score $S=\{s\}$;
        \STATE Compute importance $\alpha$ Eq.~\ref{eq:piecewise} for each image using $S$;
        
        \textbf{Second learning stage:}
        \STATE $\Phi^0 := \Phi^{K}$;
        \FOR{\textit{i}=1 \textbf{to} $K$}
        \STATE Obtain intermediate heatmap $Y_{inter}$, main heatmap $Y$, offset $O$ and size prediction $S$;
        \STATE Calculate the detection loss $\mathcal{L}^{\alpha}$ following the Eq.~\ref{eq:alphas};
        \STATE Update model weight $\Phi^i := \mathcal{A}(\mathcal{L}, \Phi^{i-1})$;
        \ENDFOR\\
        \STATE $\Phi^{*} := \Phi^{K}$;
        \RETURN $\Phi^{*}$;
    \end{algorithmic}
\end{algorithm}
\subsection{IoU Guided Sample Re-weighting}
\label{ssec:ahsm}

During the first learning stage, we noticed that the training loss of some samples did not correspond with their inference IoU scores. Specifically, some samples with higher IoU (\textit{i.e.} easy samples) still provide larger loss, as shown in the red and yellow abnormal areas in Fig.~\ref{fig:hard_mining_result}. This phenomenon may bias the training of the detection model. We speculate that this may be due to the camera-moving characteristic of the colonoscopy, resulting in video jitter and significant changes in background features. To address this, we propose an IoU-guided sample re-weighting mechanism to adaptively adjust the importance weight $\alpha$ according to the inference IoU $s$ of the sample.
\begin{equation}
    \label{eq:piecewise}
    \alpha = 1 - s
\end{equation}
where $s$ is predicted by the trained model after the first learning stage. After re-weighting, the training losses are more reasonably aligned with inference IoUs, and most samples are distributed in normal areas, as shown in Fig.~\ref{fig:hard_mining_result}.

In the second learning phase, we only optimize the detection loss $\mathcal{L}^{\alpha}$ using re-weighted $\mathcal{L}_{hm}^{inter}$, $\mathcal{L}_{hm}^{main}$, $\mathcal{L}_{o}$ and $\mathcal{L}_{s}$ with weight $\lambda$. Take the size prediction loss as an example. The re-weighted loss can be expressed as follows:
\begin{equation}
    \label{eq:alphas}
    \mathcal{L}_{s}^{\alpha} = \frac{1}{N} \sum^N_{k=1} \alpha |\mathcal{{S}}_{\tilde{p}k} - \tilde{s_{k}}| 
\end{equation}
Other losses are in the same form with $\alpha$. In summary, the whole training pipeline of our detection system is described in Alg.~\ref{alg:system}.

\begin{table}[t]
    \centering
    \caption{The colonoscopy polyp detection datasets used in our experiments.}
    \label{tab:dataset}
    \setlength{\tabcolsep}{3pt}
    \resizebox{\linewidth}{!}{\begin{tabular}{ccccc}
    \toprule[1.5pt]
    Dataset               & Train Images  & Test Images  & Input size          & Availability \\\midrule
    LHR Database-L        & 64,996        & 12,934       & 1504 $\times$ 1080  & Copyrighted   \\
    LHR Database-S        & 15,916        & 4,040        & 1082 $\times$ 940   & Copyrighted  \\
    SUN Database$^2$          & 19,544        & 12,522       & 1158 $\times$ 1008  & Public  \\
    LDPolypVideo$^3$          & 20,942        & 12,933       & 560$\times$ 480     & Public  \\
    CVC-VideoClinicDB$^4$     & 9,775         & 2,030        & 384 $\times$ 288    & Public  \\ 
    PolypGen$^5$              & 1264          & 83           & 1350 $\times$ 1080  & Public \\
    \bottomrule[1.5pt]
    \multicolumn{5}{l}{\small $^2$\text{http://sundatabase.org/}}\\
    \multicolumn{5}{l}{\small $^3$\text{https://github.com/dashishi/LDPolypVideo-Benchmark}}\\
    \multicolumn{5}{l}{\small $^4$\text{https://giana.grand-challenge.org/}}\\
    \multicolumn{5}{l}{\small $^5$\text{https://doi.org/10.7303/syn26376615}}\\
    \end{tabular}
    }
\end{table}

\begin{figure}[t]
    \centering
    \includegraphics[width=\linewidth]{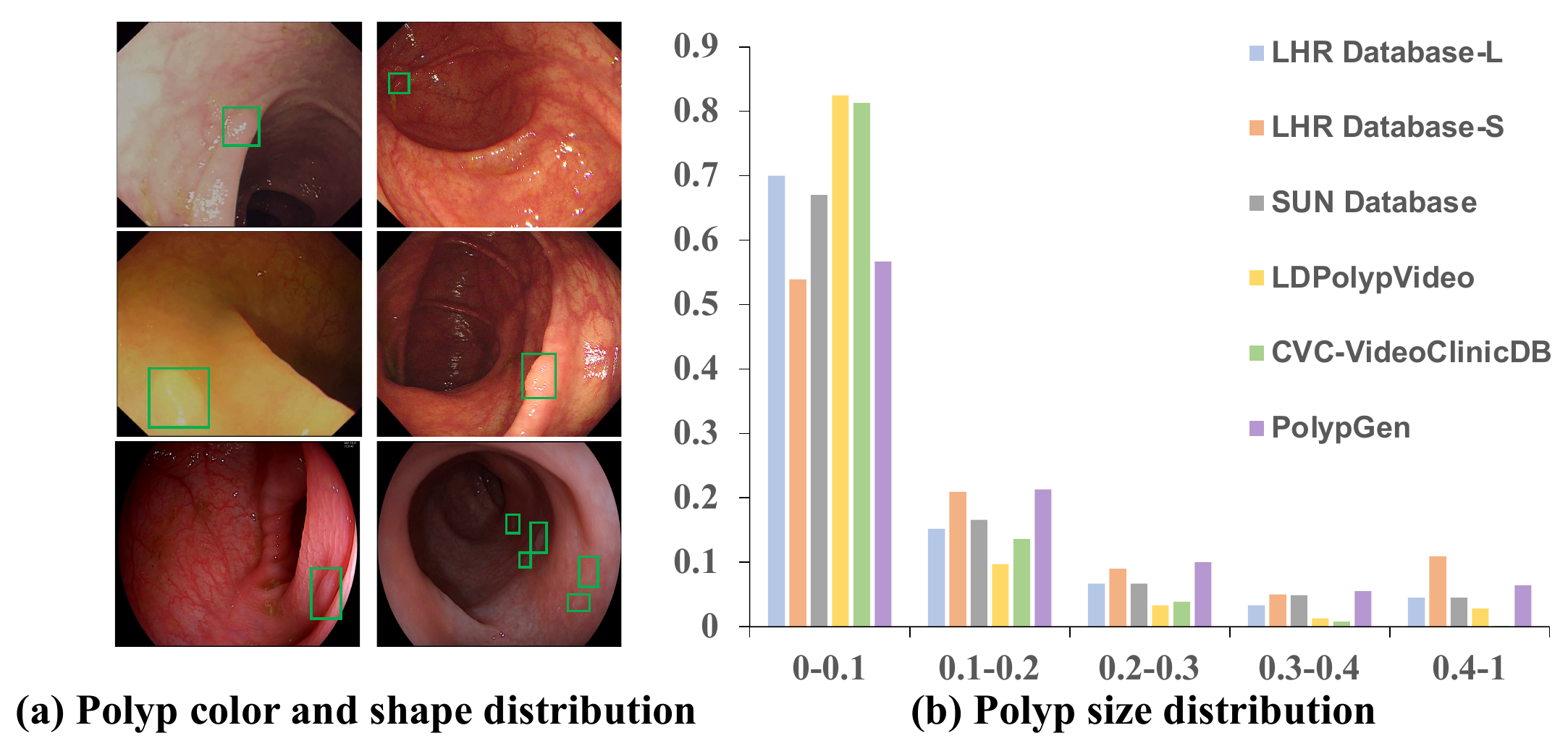}
    \caption{
    (a) Polyp samples with different shapes, sizes, and colors. 
    (b) The histogram of polyp size. The horizontal axis illustrates the proportion of the polyp area to the image area. 
    The vertical axis shows the proportion of the polyp samples of a specific size to the total samples.}
    \label{fig:data}
\end{figure}

\begin{table*}[t]
    \centering
    \caption{Quantitative comparison of different detectors and our ECC-PolypDet on SUN Database with default PVTv2 backbone. \textbf{Bold} denotes the best results. R50 denotes ResNet-50 backbone}
    \label{tab:result_public}
    \setlength{\tabcolsep}{7pt}
    \resizebox{\linewidth}{!}{\begin{tabular}{l|cccc|cccc|cccc}
    \toprule[1.5pt]
    \multirow{2}{*}{Models} & \multicolumn{4}{c|}{SUN} &\multicolumn{4}{c|}{LDPolypVideo} &\multicolumn{4}{c}{CVC-VideoClinicDB} \\ \cmidrule{2-13} 
    & AP   & P    & R    & F1   & AP   & P    & R    & F1   & AP   & P    & R    & F1 \\ \midrule
    CenterNet-R50~\cite{centernet}            & 67.3 & 74.6 & 65.4 & 69.7 & 57.3 & 70.6 & 43.8 & 54.1 & 86.5 & 92.0 & 80.5 & 85.9    \\
    DINO-R50~\cite{dino}                      & 75.6 & 81.5 & 72.3 & 76.6 & 63.3 & 68.3 & 51.1 & 58.5 & 89.7 & 93.1 & 89.2 & 91.1   \\
    STFT-R50~\cite{stft}                      & 76.0 & 81.5 & 72.4 & 76.7 & 65.1 & 72.1 & 50.4 & 59.3 & 90.5 & 91.9 & \textbf{92.0} & 91.9   \\
    \textbf{ECC-PolypDet-R50} (\textit{Ours}) & \textbf{77.8} & \textbf{81.7} & \textbf{74.7} & \textbf{78.0} & \textbf{66.2} & \textbf{74.0} & \textbf{52.4} & \textbf{61.4} & \textbf{89.8} & \textbf{92.8} & 91.4 & \textbf{92.1}  \\
    \midrule
    Faster R-CNN~\cite{faster_rcnn}           & 78.3 & 66.5 & 83.2 & 73.9 & 63.8 & 71.6 & 49.4 & 58.5 & 88.2 & 84.6 & \textbf{98.2} & 90.9     \\
    CenterNet                                 & 77.6 & 79.4 & 78.1 & 78.7 & 59.7 & 72.6 & 46.6 & 56.8 & 87.0 & 92.4 & 82.0 & 86.9    \\
    Sparse R-CNN~\cite{sparsercnn}            & 80.9 & 85.1 & 82.3 & 83.7 & 65.3 & 72.0 & 50.5 & 59.5 & 87.9 & 85.1 & 96.4 & 90.4    \\
    YOLOv5~\cite{YOLOv5}                       & 77.0 & 76.4 & 80.3 & 78.3 & 58.5 & 68.3 & 46.9 & 55.6 & 87.5 & 91.4 & 83.5 & 87.3   \\
    YOLOX~\cite{yolox}                        & 79.4 & 77.1 & 82.3 & 79.6 & 64.2 & 70.1 & 49.9 & 58.3 & 86.7 & 92.6 & 85.6 & 89.0   \\
    Deformable DETR~\cite{deformabledetr}     & 81.3 & 83.6 & 79.1 & 81.3 & 64.0 & 65.0 & 51.0 & 57.2 & 85.4 & 91.8 & 79.6 & 85.3   \\
    DINO                                      & 81.8 & 87.3 & 79.9 & 83.4 & 65.5 & 71.6 & 51.7 & 60.0 & 90.3 & \textbf{93.6} & 90.1 & 91.8   \\
    ColonSeg~\cite{colono_seg}                & 65.3 & 75.7 & 62.1 & 68.2 & 58.2 & 67.5 & 46.0 & 54.7 & 80.5 & 78.9 & 88.4 & 83.4   \\
    STFT                                      & 81.6 & 86.3 & 80.7 & 83.4 & 65.9 & 74.5 & 50.6 & 60.2 & 90.8 & 92.6 & 92.3 & 92.4   \\
    \textbf{ECC-PolypDet} (\textit{Ours})     & \textbf{82.2} & \textbf{87.7} & \textbf{84.2} & \textbf{85.8} & \textbf{68.5} & \textbf{77.1} & \textbf{53.9} & \textbf{63.4} & \textbf{91.0} & 93.3 & 92.8 & \textbf{93.0}   \\ \bottomrule[1.5pt]
    \end{tabular}
    }
\end{table*}

\section{Experiments}
\label{sec:experiment}

\subsection{Dataset}
\label{ssec:dataset}
We evaluate our method with competitors on six different challenging datasets in our experiments, as shown in Table~\ref{tab:dataset}.

\textbf{Public Dataset}
(1) SUN Colonoscopy Video Database~\cite{suncolon,sunwebsite} is a colonoscopy video dataset collected by Showa University \& Nagoya University database which includes $49,136$ polyp frames. We follow the settings in SUN-SEG~\cite{sunseg}, where SUN-SEG is re-organized by Ji~\textit{et al.} based on the SUN database. We use 112 clips for training and the rest 54 clips for testing.
(2) LDPolypVideo~\cite{ldpolypvideo} dataset is a large-scale and diverse colonoscopy video dataset. It contains $160$ colonoscopy video clips where $33,884$ frames contain at least one polyp. We split $100$ clips as the training set and 60 clips as the testing set.
(3) CVC-VideoClinicDB database~\cite{cvc_1,cvc_2}, which is composed of more than 40 sequences collected at the Hospital Clinic of Barcelona, Spain.
Following the settings in~\cite{stft}, we split 18 video clips with annotations into training sets (14 videos) and validation sets (4 videos) for evaluation.
(4) PolypGen~\cite{ali2023multi} is a large multicentre dataset collected from six unique centres. It contains $1347$ colonoscopy images from different patients and populations. To test the generalization ability of our method, we follow the dataset instructions and take the first five centres as the train set and the sixth centre as the test set.

\textbf{Private Dataset}
In order to provide more substantial research resources for automatic polyp diagnosis and further test the generalization ability of our model in real-world scenarios, we collect a large-scale, high-quality, and real-world colonoscopy video database that contains a variety of polyps and more complex colon environments, namely LHR Database. They are identified as LHR Database-L (Large) and S (Small) based on the image sizes.
(1) LHR Database-L provides 300 video clips with 93,876 frames, at least 83,605 frames of which contain one polyp. We split 16\% of video clips as the test set and the rest as the training set.
(2) LHR Database-S contains 152 video clips, 80\% of them contain one polyp. We split 20\% of video clips as the test set and the rest as the training set.

\textbf{Challenges}
Those five datasets we chose satisfied the challenges we mentioned in section.~\ref{sec:introduction}. As shown in Fig.~\ref{fig:data} (a), some typical polyp samples taken from those six datasets demonstrate the concealed property of polyps in most colonoscopy videos. We also analyzed the polyp size distribution in all datasets. As shown in Fig.~\ref{fig:data} (b), the vast majority of polyps have a size that is less than 0.1\% of the total image area. 

\begin{table*}[t]       
    \centering
    \caption{Quantitative comparison of different detectors and our ECC-PolypDet on LHR Database-L and LHR Database-S. $T_{train}$: training time; FPS: testing frame per second.}
    \label{tab:result_private}
    \setlength{\tabcolsep}{9pt}
    \resizebox{\linewidth}{!}{\begin{tabular}{l|cccc|cccccc}
    \toprule[1.5pt]
    \multirow{2}{*}{Models} & \multicolumn{4}{c|}{LHR Database-L} &\multicolumn{6}{c}{LHR Database-S} \\ \cmidrule(lr){2-11}   
                    &  AP   & P    & R    & F1   & AP   & P    & R    & F1   & $T_{train}$ (s) & FPS    \\ \midrule
    Faster R-CNN    &  37.2 & 51.4 & 45.3 & 48.2 & 68.7 & 67.8 & 76.6 & 71.9 & 10201.3     & 45.3       \\
    CenterNet       &  48.3 & 70.7 & 53.1 & 60.6 & 71.0 & 77.8 & 74.4 & 76.1 & 8524.5      & 51.5 \\
    Sparse R-CNN    &  48.1 & 68.5 & 53.5 & 60.1 & 76.0 & 84.3 & 76.2 & 80.0 & 12622.1     & 40.2 \\
    YOLOv5          &  47.7 & 68.2 & 48.1 & 56.4 & 68.9 & 82.4 & 64.7 & 72.5 & 7320.6   & 54.3 \\
    YOLOX           &  50.8 & \textbf{75.4} & 50.8 & 60.7 & 70.9 & 85.2 & 66.2 & 74.5 & 9002.5 & 47.1 \\
    Deformable DETR &  49.7 & 58.5 & \textbf{55.8} & 57.1 & 71.8 & 79.4 & 74.9 & 77.1 & 15962.2& 26.4 \\
    DINO            &  51.8 & 73.5 & 52.4 & 61.1 & 76.5 & 80.6 & 79.9 & 80.3 & 21247.3     & 20.5 \\
    ColonSeg        &  36.6 & 64.4 & 48.6 & 55.4 & 63.5 & 66.3 & 69.3 & 67.7 & 10504.1     & 44.1 \\
    STFT            &  50.2 & 70.9 & 51.6 & 59.7 & 76.1 & 83.6 & 78.3 & 80.8 & 34688.7     & 8.5 \\
    \textbf{ECC-PolypDet} (\textit{Ours}) & \textbf{54.3} & 74.0 & 55.1 & \textbf{63.2} & \textbf{79.3} & \textbf{86.2} & \textbf{83.0} & \textbf{84.6}  & 9902.6 & 46.1 \\ \bottomrule[1.5pt]
    \end{tabular}
    }
\end{table*}

\subsection{Implementation Details}

\subsubsection{Evaluation Metrics}

We use the standard metrics presented in the \textit{MICCAI 2015 Automatic Polyp Detection} for a fair comparison of all methods, including precision, recall, F1-score, and AP. Besides, we also employ the widely used metrics in object detection for an objective evaluation.

The \textit{Precision} (P) measures the rate of the predicted positive samples that are true positive. Higher precision can more effectively prevent the false alarm. The \textit{Recall} (R) represents the proportion of the true positives that are correctly classified. Higher recall ensures more polyps can be diagnosed:
\begin{equation*}
    \begin{aligned}
         \mathrm{P}=\frac{TP}{TP+FP}, \quad\mathrm{R}=\frac{TP}{TP+FN},
    \end{aligned}
\end{equation*}

\noindent where \textit{TP}, \textit{FP}, \textit{TN}, and \textit{FN} represent true positives, false positives, true negatives, and false negatives, respectively.

The \textit{F1-score} (F1) calculates the harmonic weight of the precision and recall.

\begin{equation*}
    \begin{aligned}
        \mathrm{F1}&=\frac{2P\cdot R}{P+R}
    \end{aligned}
\end{equation*}

For clinical applications, a low false positive rate (P) and a low false negative rate (R) are equally important. The precision and recall alone cannot fully reflect the performance of the model. Therefore, in this paper, we pay more attention to F1, which can evaluate the performance more comprehensively.

\textit{The Average Precision} (AP) is the area under the curve (AUC) of the Precision $\times$ Recall curve. Following the standard of PASCAL VOC challenge, AP is obtained by interpolating the precision at all levels of recall between 0 and 1:

\begin{equation*}
    \begin{aligned}
        &\mathrm{AP} = \sum_{n=0}(r_{n+1}-r_n)\rho_{interp}(r_{n+1}) \\
    &\rho_{interp}(r_{n+1})=\mathop{max}\limits_{\Tilde{r}: \Tilde{r}\geqslant r_{n+1}} \rho(\Tilde{r})
    \end{aligned}
\end{equation*}

\noindent where $\rho(\Tilde{r})$ is the measured precision at recall $\Tilde{r}$.

\subsubsection{Experimental Setup and Configuration}

We chose nine well-known/SOTA object detection methods to compare with our ECC-PolypDet. We implement our model based on PyTorch~\cite{torch}. The object detection competitors are implemented based on MMDetection~\cite{mmdet}. Other polyp detection methods are re-implemented based on their open-source code. We select the conventional CNN model ResNet-50\cite{resnet} and SOTA transformer model PVTv2~\cite{pvt_v2} as the feature extractor (backbone) where their weights are pre-trained on ImageNet~\cite{imagenet}. It is worth noting that the backbone network can be replaced by any mainstream neural network. All the methods are trained on a single NVIDIA A100 GPU. We set the number of intermediate layers $L=1$ and the batch size $N=16$. During training, we randomly crop and resize the images to $ 512 \times 512 $ and normalize them using ImageNet settings. Random rotation and flip are used for data augmentation. Our method is trained using the Adam optimizer with cosine annealing weight decay for 20 epochs. The initial learning rate is set to 0.0001. 

\begin{table}[t]
    \centering
    \caption{Comparison of the generalization capability of different detectors and our ECC-PolypDet on LHR Database training set and SUN Hard test set.}
    \label{tab:result_generalization}
    \setlength{\tabcolsep}{10pt}
    \resizebox{\linewidth}{!}{\begin{tabular}{l|cccc}
    \toprule[1.5pt]
    \multirow{2}{*}{Models} & \multicolumn{2}{c}{LHR $\rightarrow$ SUN} & \multicolumn{2}{c}{PolypGen} \\ \cmidrule{2-5}
                    & AP   & F1   & AP   & F1  \\ \midrule
    Faster R-CNN    & 79.6 & 77.0 & 64.1 & 72.4    \\
    CenterNet       & 75.8 & 77.4 & 67.5 & 74.8    \\
    Sparse R-CNN    & 81.2 & 80.1 & 69.6 & 77.2    \\
    YOLOv5          & 76.7 & 73.1 & 66.1 & 72.8   \\
    YOLOX           & 76.7 & 73.1 & 68.3 & 76.0   \\
    Deformable DETR & 80.1 & 79.3 & 60.3 & 67.7   \\
    DINO            & 80.9 & 80.3 & 67.2 & 75.2   \\
    ColonSeg        & 62.0 & 67.9 & 62.6 & 70.2   \\
    STFT            & 78.3 & 77.1 & 70.4 & 78.9   \\
    \textbf{ECC-PolypDet} (\textit{Ours})     & \textbf{83.2} & \textbf{84.3} & \textbf{75.5} & \textbf{82.2} \\ \bottomrule[1.5pt]
    \end{tabular}}
\end{table}

\subsection{Quantitative Comparison}
\subsubsection{Learning Ability}
To evaluate the learning ability of our model, we first trained and tested it on five datasets, respectively. The results are shown in Table.~\ref{tab:result_public} and Table.~\ref{tab:result_private}. In Table.~\ref{tab:result_public}, The top part shows the results of our model with other polyp detectors based on the ResNet-50 backbone, and the bottom part exhibits the results based on the PVTv2 backbone. It demonstrates that our proposed ECC-PolypDet is superior to other methods of polyp detection, and the result is robust on different backbone networks. Specifically, compared with the CenterNet baseline, our ECC-PolypDet achieves a significant improvement by $8.3\%$ F1-score with ResNet-50 and more than a $7.1\%$ with PVTv2 on the SUN database. Similar results were obtained for the LDPolypVideo and CVC-VideoClinicDB datasets. Moreover, ECC-PolypDet outperforms the second-best polyp detector STFT by $0.6\%$ AP and $2.4\%$ F1-score. In Table.~\ref{tab:result_private}, ECC-PolypDet achieves robust results on both LHR L and S database, with more than a $4.1\%$ AP and $3.8\%$ F1-score gain compared with STFT. In addition, ECC-PolypDet also achieves competitive training and inference speed, taking $9902.6$ seconds for training on the LHR-S database, and the testing FPS is $46.1$, which is slightly slower than the lightweight models CenterNet and YOLOv5.
Although ECC-PolypDet does not offer the same real-time speed as YOLOv5, it provides much higher accuracy.

\begin{figure*}[t]
    \centering
    \includegraphics[width=0.9\linewidth]{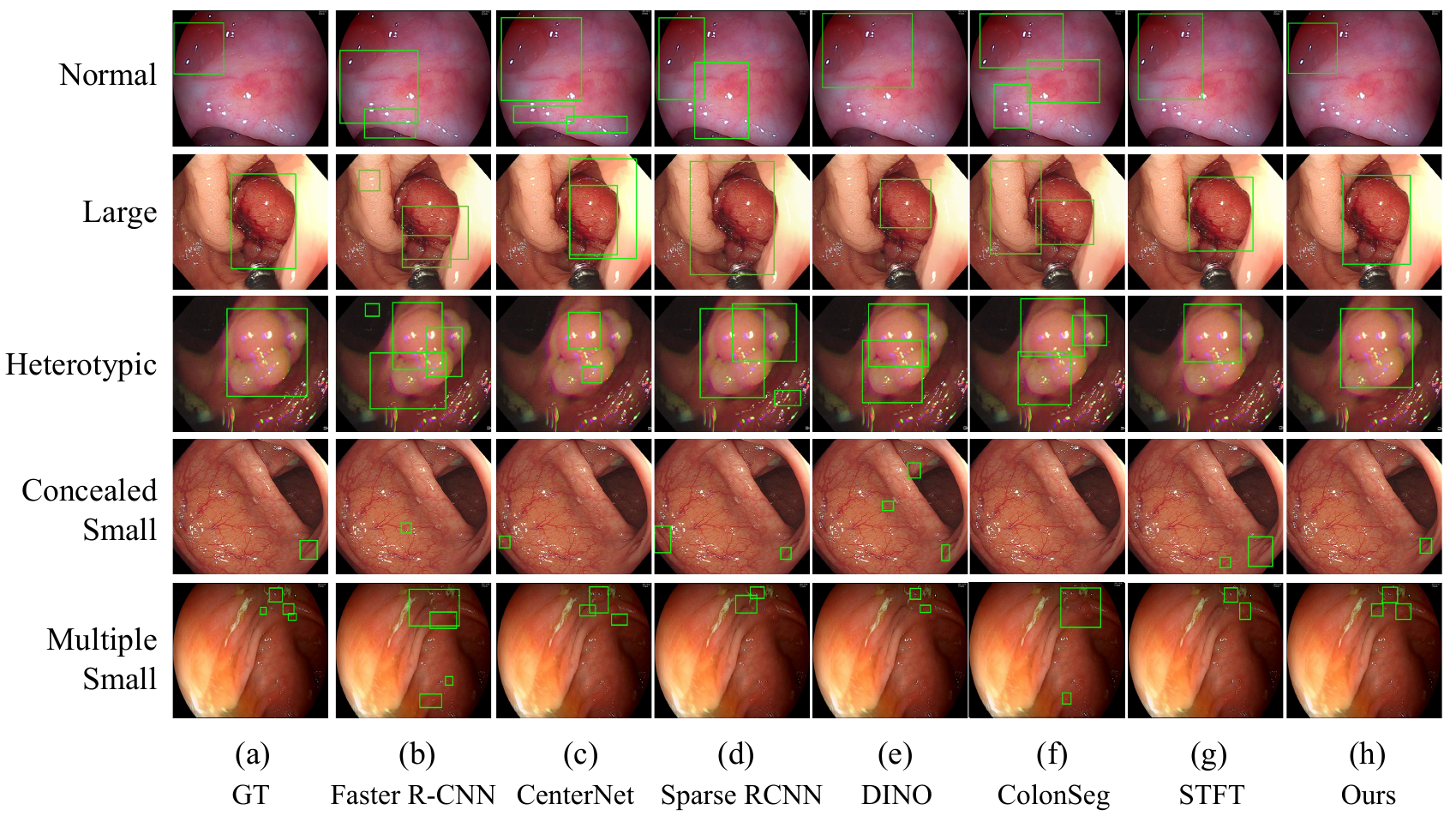}
    \caption{Qualitative visualization comparison between different detectors and ECC-PolypDet.}
    \label{fig:qualitative}
\end{figure*}

\subsubsection{Generalization Capability}
We conduct cross-domain experiments to evaluate the generalization capability of our method and other competitors. First, we merge the LHR Database-L and LHR Database-S to form a large-scale training set (LHR Database), which contains 83,858 polyp frames in total. Then we evaluate the trained models on the SUN Database test set, in which the testing data significantly differs from the the LHR training data in its essence. As shown in the first two columns of Table.~\ref{tab:result_generalization}, ECC-PolypDet shows better generalization ability compared with other methods and obtains a F1-score of $84.3\%$ which is $4.0\%$ higher than the second-best method DINO. Second, we perform another experiment on the PolypGen dataset, which consists of colonoscopy data from six different centres. It is an ideal material to test the generalization capability of methods. We train the models on data from the first five centres (C1-C5), and report the testing score on the last centre (C6). According to the last two columns of Table.~\ref{tab:result_generalization}, ECC-PolypDet achieves the highest AP and F1-score using the unique supervised contrastive learning to distinguish polyp features. Notably, it can be observed that some methods drop dramatically due to the domain gap. \textit{i.e.} STFT from $83.4$ to $77.1$ on SUN database, which may be caused by the lack of ability to extract and distinguish confusing features from generic features. The state-of-the-art performance demonstrates the robustness and general applicability of our proposed ECC-PolypDet.

\subsection{Qualitative Comparison}

\begin{figure}[t]
    \centering
    \includegraphics[width=\linewidth]{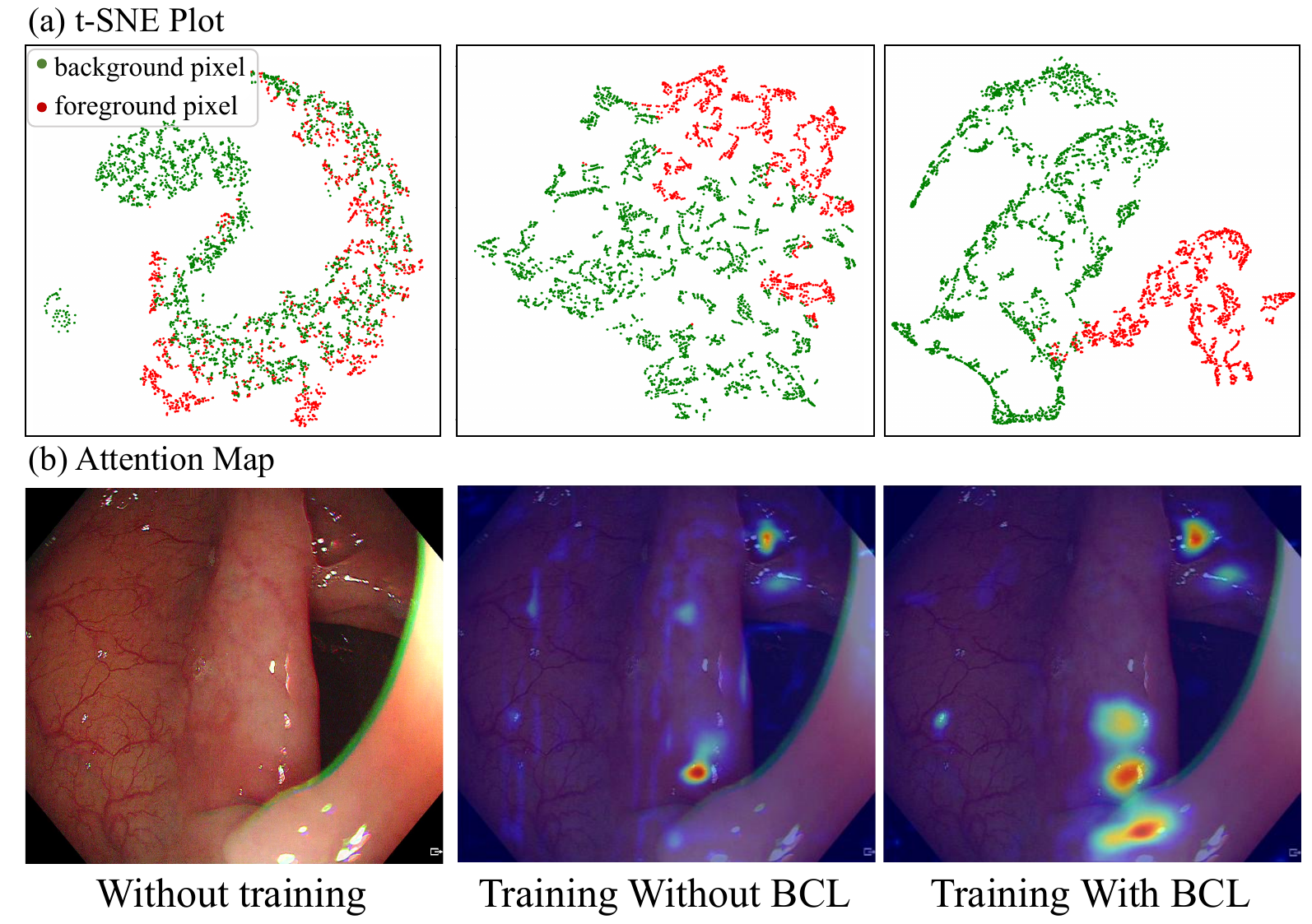}
    \caption{t-SNE plots for illustration of embedding space with BCL (Top). Attention maps for hard samples with concealed polyps (Bottom).}
    \label{fig:sne}
\end{figure}

Fig.~\ref{fig:qualitative} depicts the qualitative results of different detectors on five colonoscopy samples. We select five cases from two aspects: normal size ($ 1^{st}-3^{rd} $ rows) with heterotypic polyps ($ 3^{rd} $ row), small size ($ 4^{rd}- 5^{th} $ rows) with multiple small polyps($ 5^{th} $ row). It can be observed that heterotypic polyps are hard to detect since they are easily misidentified as multiple polyps. Similarly, multiple small polyps may be overlooked or incorrectly detected as a single polyp. In contrast, ECC-PolypDet can accurately predict the boxes, reducing the false positives and producing more reliable results. Fig.~\ref{fig:sne} illustrates the impact of BCL on recognition ability. The t-distributed stochastic neighbor embedding (t-SNE) plot shows the results on embedding space (Fig.~\ref{fig:sne} (a)). After supervised contrastive learning, feature embeddings of different classes can be well separated, which is the basis for generating accurate detection results. As for the attention map (Fig.~\ref{fig:sne} (b)), the baseline method focuses on the wrong location, while our proposed method can accurately identify the concealed polyp area and shape. Overall, these results demonstrate the effectiveness of our method in detecting polyps of different sizes and types.

\subsection{Detection Results of Different Sizes}

To verify the robustness of our method in handling polyps of various sizes, we further conducted experiments and evaluated the F1-score of different methods across different polyp size ranges. As depicted in Table.~\ref{tab:small-polyp-detection}, we divide the polyp images into five groups, where in each group, the proportion range of polyp size to image size is: $\leq 10\%$, $10\% \sim 20\%$, $20\% \sim 30\%$, $30\% \sim 40\%$, $\geq 40\%$. All the results are tested on SUN Database test sets under the same settings. It can be observed that ECC-PolypDet achieves the best results in almost all sizes except the result of $>40\%$, which is slightly lower than STFT. We suspect that relatively large polyps may cause the model to confuse them with the background area. Nevertheless, these results demonstrate the robustness of our ECC-PolypDet in detecting polyps of various sizes.

\subsection{Ablation Studies}

\subsubsection{Analysis of the effectiveness of each component}
For the effectiveness of the design of each component, we conducted the ablation experiments by adding four components step by step. Table.~\ref{tab:ablation-modules} summarizes the results. On the SUN testing set, the baseline CenterNet with PVTv2 achieves an F1-score of $ 78.7\% $. Our method obtains a significant performance gain by adding the proposed BCL strategy, which improves the F1-score by $ 3.0\% $, indicating the effectiveness of contrastive information learning. Fig.~\ref{fig:sne} shows the t-SNE~\cite{t_sne} visualization of fused backbone feature $F$ from two classes. Compared with other settings, our box-assisted contrastive learning can enhance a much better intra-class compactness and inter-class discrepancy. The SFFPN further gains a $2.6\%$ F1-score improvement by reducing the severe misalignment in the downsampling and upsampling path. Moreover, replacing the CenterNet with our HP further improves the F1-score by $ 3.3\% $ due to the enhancement of the model's capability of capturing information on small polyps. Finally, the ISR strategy boosts the performance by $ 2.0\% $ F1-score, which shows the impact of mining hard samples for detection models. Notably, the ISR contributes more to the recall score, indicating the effectiveness of this strategy for mining small polyps and polyps with low color/texture contrast. In addition, the effectiveness of each component is shown in Table.~\ref{tab:small-polyp-detection}. Combining all the components, our ECC-PolypDet achieves an improvement of $7.1\%$ on the F1-score, which demonstrates the effectiveness of our design.

\begin{table}[t]
    \centering
    \caption{F1-score Performance of various methods and modules across different polyp sizes. *CenterNet is the baseline model}
    \label{tab:small-polyp-detection}
    \setlength{\tabcolsep}{4pt}
    \resizebox{\linewidth}{!}{\begin{tabular}{c|ccccc}
    \toprule[1.5pt]
    \multirow{2}{*}{Method} &  \multicolumn{5}{c}{Relative polyp size ratio}\\\cmidrule{2-6}  & $<10\%$   & $10\%\sim20\%$      & $20\%\sim30\%$     & $30\%\sim40\%$    & $>40\%$ \\ \midrule
    CenterNet* &  68.9  & 76.3 & 70.5 & 60.3 & 67.0 \\
    DINO   &  65.4  & 74.9 & 72.6 & 58.8 & 65.5 \\
    STFT   &  72.5  & 77.7 & 71.0 & 62.5 & \textbf{70.4} \\
    Ours   &  \textbf{74.7}  & \textbf{82.0} & \textbf{73.8} & \textbf{63.7} & 68.7 \\ \midrule
    Baseline* &  68.9  & 76.3 & 70.5 & 60.3 & 67.0 \\
    ISR &  69.5  & 77.2 & 71.0 & 61.9 & 67.4 \\
    HP &  71.1  & 78.0 & 71.3 & 62.2 & 68.2 \\
    SFFPN &  70.8  & 78.0 & 70.7 & \textbf{64.0} & 68.0 \\
    BCL &  \textbf{72.0}  & \textbf{78.2} & \textbf{72.2} & 63.4 & \textbf{68.5} \\
    \bottomrule[1.5pt]
    \end{tabular}
    }
\end{table}

\begin{table}[t]
    \centering
    \caption{The effectiveness of proposed modules on SUN Hard test sets.}
    \label{tab:ablation-modules}
    \setlength{\tabcolsep}{3.5pt}
    \resizebox{\linewidth}{!}{\begin{tabular}{cccc|cccc}
    \toprule[1.5pt]
    BCL  &    SFFPN    &     HP         & ISR       & AP   & P     & R     & F1   \\ \midrule
    \multicolumn{4}{c|}{Baseline}        & 77.6 & 79.4 & 78.1 & 78.7 \\ 
    \checkmark &       &      &            & 79.5$_{\textcolor{green}{\uparrow}1.9}$ & 84.8$_{\textcolor{green}{\uparrow}5.4}$ & 79.6$_{\textcolor{green}{\uparrow}1.5}$ & 82.1$_{\textcolor{green}{\uparrow}3.4}$ \\
     &  \checkmark    &      &            & 79.2$_{\textcolor{green}{\uparrow}1.6}$ & 79.3$_{\textcolor{red}{\downarrow}0.1}$ & 83.5$_{\textcolor{green}{\uparrow}5.4}$ & 81.3$_{\textcolor{green}{\uparrow}2.6}$ \\
     &            & \checkmark &          & 78.8$_{\textcolor{green}{\uparrow}1.2}$ & 83.1$_{\textcolor{green}{\uparrow}3.7}$ & 78.6$_{\textcolor{green}{\uparrow}0.5}$ & 80.8$_{\textcolor{green}{\uparrow}2.1}$ \\
     &            &  &    \checkmark      & 78.2$_{\textcolor{green}{\uparrow}0.6}$ & 81.5$_{\textcolor{green}{\uparrow}2.1}$ & 78.3$_{\textcolor{green}{\uparrow}0.2}$ & 79.9$_{\textcolor{green}{\uparrow}1.2}$ \\
    \checkmark & \checkmark &    &        & 80.5$_{\textcolor{green}{\uparrow}2.9}$ & 85.6$_{\textcolor{green}{\uparrow}6.2}$ & 82.0$_{\textcolor{green}{\uparrow}3.9}$ & 83.8$_{\textcolor{green}{\uparrow}5.1}$ \\  
    \checkmark & \checkmark &  \checkmark  &        & 81.0$_{\textcolor{green}{\uparrow}3.4}$ & 87.1$_{\textcolor{green}{\uparrow}7.7}$ & 81.1$_{\textcolor{green}{\uparrow}3.0}$ & 84.0$_{\textcolor{green}{\uparrow}5.3}$ \\ 
    \checkmark & \checkmark &\checkmark& \checkmark & \textbf{82.2}$_{\textcolor{green}{\uparrow}\mathbf{4.6}}$ & \textbf{87.7}$_{\textcolor{green}{\uparrow}\mathbf{8.3}}$ & \textbf{84.2}$_{\textcolor{green}{\uparrow}\mathbf{6.1}}$ & \textbf{85.8}$_{\textcolor{green}{\uparrow}\mathbf{7.1}}$ \\ \bottomrule[1.5pt]
    \end{tabular}
    }
\end{table}

\subsubsection{Analysis of the number of intermediate stages}
We conducted an ablation study to investigate the impact of the number of stages in our proposed HP. The results are presented in Table.~\ref{tab:ablation-stage}, where the detector with 1-stage represents the original CenterNet. As shown in the table, adding one more stage (detector with 2-stage) significantly improves the performance of the baseline detector. However, adding the fourth stage leads to a significant drop in performance. Moreover, the training GPU consumption linearly increases with the number of stages. Based on these findings, we choose the 2-stage detector as the default, as it achieves a better trade-off between performance and efficiency.

\begin{table}[t]
    \centering
    \caption{The effects of the number of stages. $ \overline{1 \sim k} $ denotes the average results of all stages.}
    \label{tab:ablation-stage}
    \setlength{\tabcolsep}{3.5pt}
    \resizebox{\linewidth}{!}{\begin{tabular}{c|c|cccccc}
    \toprule[1.5pt]
    \# stage   &  test stage  & AP   & P      & R     & F1    & GPU (G) & FPS\\ \midrule
        1      &  1           & 79.2 & 84.8  & 79.6 & 82.1  & 8.1 & 51.5         \\ 
        2      & $ \overline{1\sim2} $& 82.2 & 87.7  & 84.2 & 85.8 & 11.0 & 46.1        \\
        3      & $ \overline{1\sim3} $& \textbf{82.6} & \textbf{87.5}  & \textbf{84.6} & \textbf{86.0} & 13.0 & 42.6       \\
        4      & $ \overline{1\sim4} $& 80.2 & 86.7  & 81.7 & 84.1 & 15.2  & 39.4       \\ \bottomrule[1.5pt]
    \end{tabular}
    }
\end{table}

\subsubsection{Ablation study of the loss coefficient}
\begin{figure}[t!]
    \centering
    \includegraphics[width=0.55\linewidth]{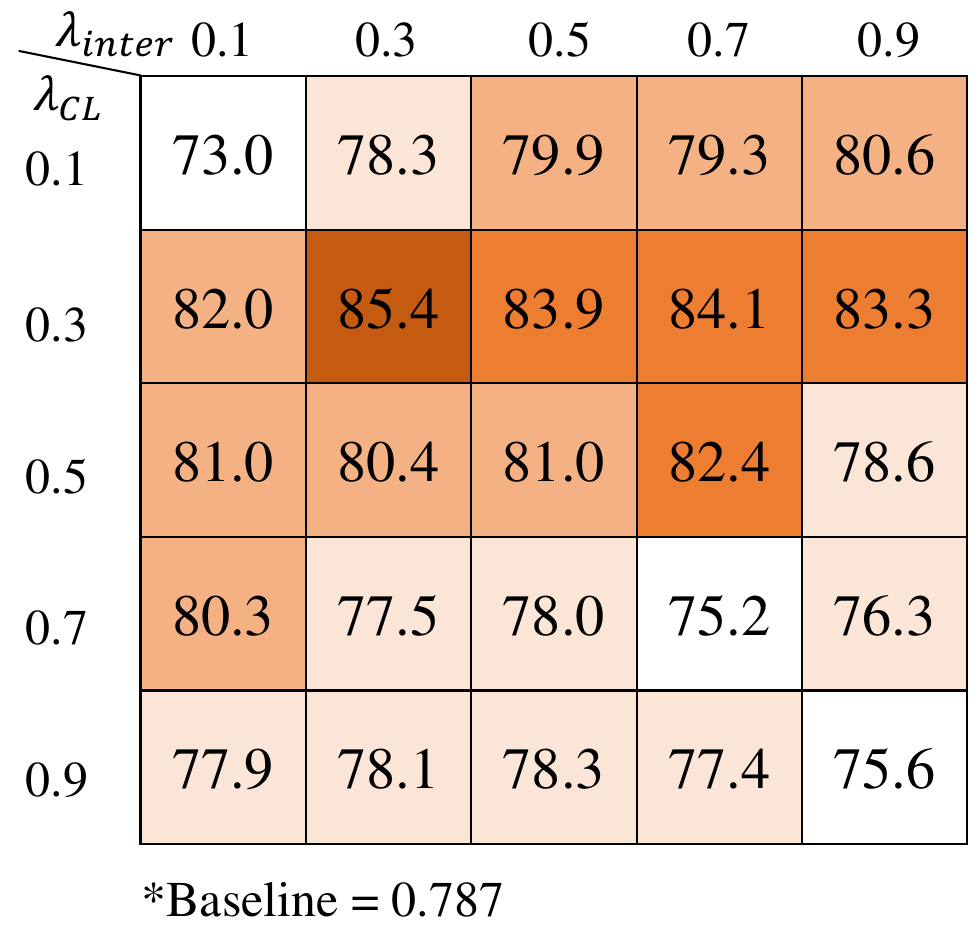}
    \caption{Detection performance (F1 score) under different loss coefficients.}
    \label{fig:loss_weight}
\end{figure}

We conducted experiments to investigate the effects of the two additional loss components $ \lambda_{CL} $ and $ \lambda_{inter} $ introduced in ECC-PolypDet. Specifically, we gradually varied the values of $ \lambda_{CL} $ and $ \lambda_{inter} $ and measured the resulting impact on the model's performance. Fig.~\ref{fig:loss_weight} demonstrates that the F1-score improves with increasing values of $ \lambda_{CL} $ and $ \lambda_{inter} $. 
However, the model's performance degrades rapidly once these values exceed a certain threshold (\textit{e.g.}, $0.5$). Based on these findings, we set both $ \lambda_{CL} $ and $ \lambda_{inter} $ as the default value of $0.3$.

\section{Discussion}
In recent years, the extensive application of deep convolutional neural networks (CNNs) in medical image analysis has yielded significant breakthroughs across various domains. These architectures, initially developed for natural image processing, require adaptation and fine-tuning for effective use in medical image analysis. Our primary objective is to identify the most suitable framework for colonoscopy video. Two-stage or transformer-based methods like Faster R-CNN or DINO exhibit exceptional object detection capabilities but are susceptible to noticeable delays during testing due to their computationally intensive nature. Conversely, one-stage methods such as YOLO or CenterNet excel in real-time inference, albeit with a marginal compromise on performance. In the context of colonoscopy, the speed of inference holds paramount importance. Among real-time frameworks, CenterNet stands out. In comparison to YOLO, it eliminates the need for anchor boxes and demonstrates robustness in handling variations in polyp shapes arising from diverse viewing angles. Furthermore, CenterNet employs a straightforward keypoint localization approach, which we consider the optimal choice for the task of polyp object detection. During the design process, we maintained CenterNet's real-time capabilities and introduced additional computations only during training to strike the finest balance between speed and accuracy.

After consulting with colonoscopy clinicians, we discovered that unstable video quality resulting from internal artifacts, such as concealed polyps, bubbles, fecal remnants, and water flow, will interfere with the observation of colonic mucosa and lead to incorrect judgments by doctors. Unfortunately, these are also obstacles for existing polyp detectors and significantly hinder the automatic polyp detection performance. We have considered improving our method's performance involves adding more challenging samples. However, this requires additional specialized data and paired labels. Therefore, we proposed a two-stage training strategy, adjusting the weights in the second stage to focus on difficult samples. Additionally, during training, we incorporated a lightweight contrastive learning module to reduce the impact of indistinguishable polyps by learning the relationship between polyp regions and the background.

Our algorithm has been successfully integrated into automated colonoscopy detection assistance software and is currently undergoing clinical testing. This powerful system has the ability to process colonoscopy videos both online and offline, providing accurate and timely test results. The benefits of this technology are manifold: it allows doctors to quickly and easily identify polyps during clinical diagnosis, thereby reducing the rate of missed diagnoses. Moreover, it offers a valuable platform for training new physicians in the field of colonoscopy operations, ensuring that they receive the best possible guidance and support.

\begin{figure}[t]
    \centering
    \includegraphics[width=0.9\linewidth]{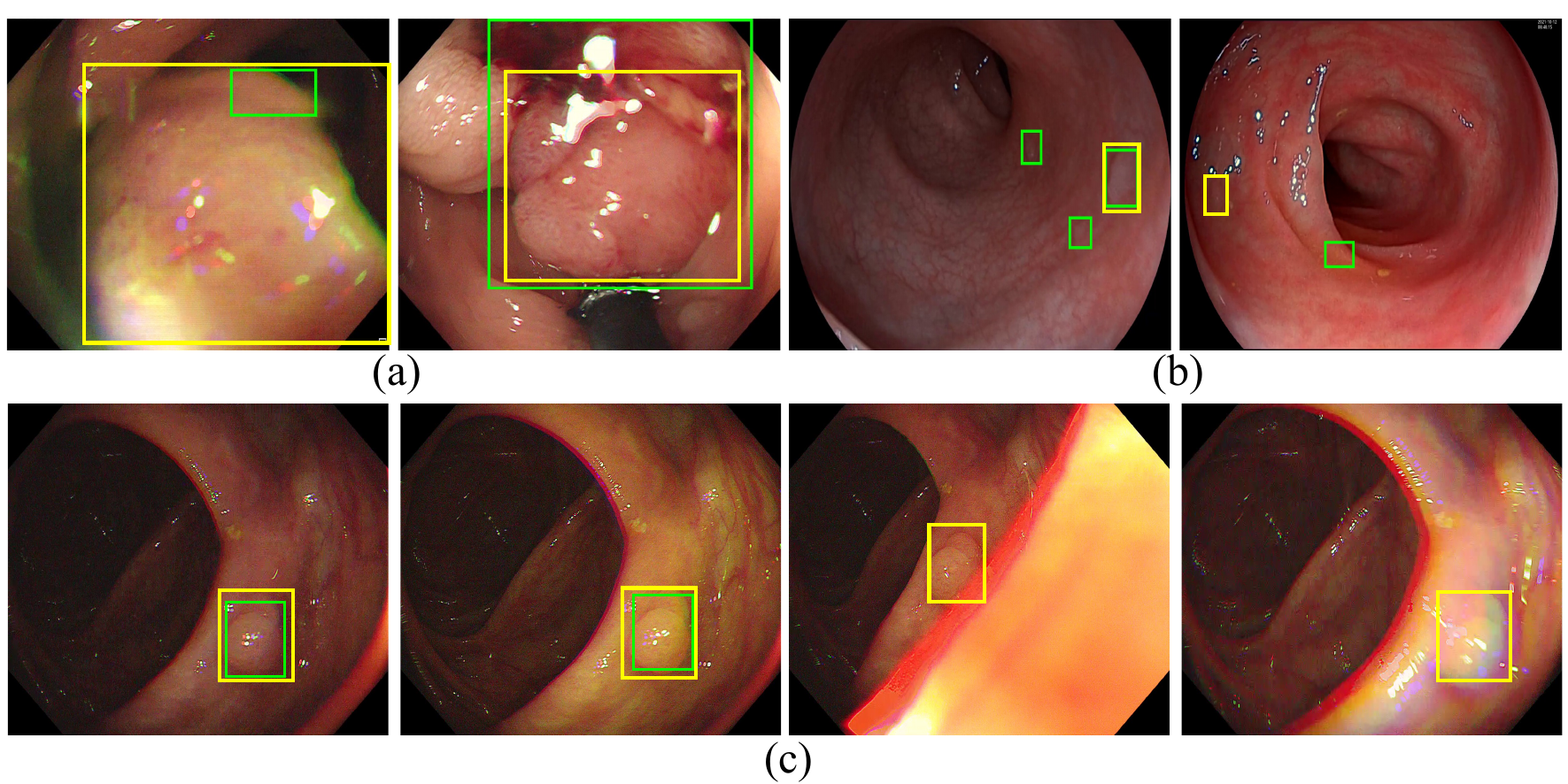}
    \caption{Illustration of failure cases. (a) False positives with large or distorted shapes polyps. (b) False positives with ambiguous boundary polyps. (c): False negatives in sequential data. The yellow box denotes the ground truth label, and the green box denotes predictions}
    \label{fig:failure_case}
\end{figure}

Despite the promising detection performance achieved by our method across various datasets, it is important to acknowledge that the model may yield suboptimal results under specific conditions. Fig.~\ref{fig:failure_case} illustrates the examples of the failure cases. One notable challenge arises when our methods encounter polyps with large sizes and distorted shapes (Fig.~\ref{fig:failure_case} (a)). The possible reason is that polyps undergo deformation due to the rapid camera movement, causing a misalignment between the current and adjacent foreground features. Another challenge arises in a continuous video segment where issues like image blurriness and partial occlusion can confuse the model and divert its focus from the intended targets, resulting in instances of false negatives (Fig.~\ref{fig:failure_case} (c)). Additionally, our model struggles when it comes to locating polyps with unclear or ambiguous boundaries (Fig.~\ref{fig:failure_case} (b)), which are similar in color and shape to the intestinal border, making them concealed within the background. This is the fundamental challenge of polyp detection and needs further research.

We believe that there are several promising directions for further improvement in the future:
\textbf{Temporal Information Integration}: Fully leveraging the temporal coherence by collaborating multiple frames can enhance the model's ability to handle dynamic scenes and improve the robustness of moving objects. This can involve techniques like frame fusion or recurrent mechanisms that allow spatiotemporal information interaction between neighboring frames.
\textbf{Global Attention Mechanism}: Adapting transformer-based architectures with global attention mechanisms is another avenue for enhancing polyp detection. These models can capture long-range dependencies and consider the entire context of an image, which is crucial when dealing with concealed polyps within complex backgrounds. Global attention can help the model focus on relevant regions and suppress irrelevant distractions, ultimately improving detection accuracy.

\section{Conclusion}
\label{sec:conclusion}
The accurate detection of polyps poses a significant challenge due to the small size of polyps and poor contrast between polyps and their surrounding tissues. In this paper, we propose the ECC-PolypDet, a fast and robust polyp detection framework that addresses these challenges through a two-stage training pipeline. In the first learning stage, we leverage box annotation to generate contrastive information and improve the feature space, thereby increasing the recall of concealed polyps. For small polyp samples, we use semantic flow to guide feature aggregation and reduce information loss, and we introduce the heatmap propagation module to enhance the heatmap in detection layers for more accurate predictions. In the second learning stage, we propose an IoU-guided sample re-weighting mechanism that scores the importance of weight to address the gap between IoU and loss during training. Our experiments on five distinct polyp detection datasets demonstrate the superior performance of ECC-PolypDet compared to other state-of-the-art detectors.

\footnotesize
\bibliographystyle{IEEEtran}
\bibliography{reference}

\end{document}

%% file: defs.tex
\def\0{{\bf 0}}
\def\1{{\bf 1}}